# The Shaky Foundations of Clinical Foundation Models:
## A Survey of Large Language Models and Foundation Models for EMRs

Michael Wornow^[1], Yizhe Xu[2], Rahul Thapa[2], Birju Patel[2], Ethan Steinberg[1], Scott Fleming[2], Michael A. Pfeffer[25], Jason Fries[2], Nigam H. Shah[2345]

[1] Department of Computer Science, Stanford University, Stanford, CA, USA
[2] Center for Biomedical Informatics Research, Stanford University School of Medicine, Stanford, CA, USA
[3] Department of Medicine, Stanford University School of Medicine, Stanford, CA, USA
[4] Clinical Excellence Research Center, Stanford University School of Medicine, Stanford, CA, USA
[5] Technology and Digital Services, Stanford Health Care, Palo Alto, CA, USA

^ corresponding author (mwornow@stanford.edu)

## Abstract

The successes of foundation models such as ChatGPT and AlphaFold have spurred significant interest in building similar models for electronic medical records (EMRs) to improve patient care and hospital operations. However, recent hype has obscured critical gaps in our understanding of these models' capabilities. We review over 80 foundation models trained on non-imaging EMR data (i.e. clinical text and/or structured data) and create a taxonomy delineating their architectures, training data, and potential use cases. We find that most models are trained on small, narrowly-scoped clinical datasets (e.g. MIMIC-III) or broad, public biomedical corpora (e.g. PubMed) and are evaluated on tasks that do not provide meaningful insights on their usefulness to health systems. In light of these findings, we propose an improved evaluation framework for measuring the benefits of clinical foundation models that is more closely grounded to metrics that matter in healthcare.

## 1. Introduction

Foundation models (FMs) are machine learning models capable of performing many different tasks after being trained on large, typically unlabeled datasets [1]. FMs have received significant attention given their impressive range of capabilities across multiple domains, from text generation [2] and video editing [3] to protein folding [4] and robotics [5].

One of the most popular FMs has been OpenAI's ChatGPT, which surpassed 100 million users within two months of release [6]. ChatGPT is a large language model (LLM), a type of FM which ingests text and outputs text in response. Though ChatGPT was trained to simply predict the next word in a sentence – it is basically an advanced autocomplete – incredible capabilities "emerged" from this training setup which allow the model to perform a wide variety of complex tasks involving language [7]. Physicians were quick to apply the model to pass medical licensing exams [8–11], simplify radiology reports [12], and write research articles [13]. In addition to text, FMs built on structured EMR data have shown the ability to predict the risk of 30-day readmission [14], select future treatments [15], and diagnose rare diseases [16].

The breakneck progress of AI over the past year has made it difficult for healthcare technology professionals and decision makers to accurately assess the strengths and limitations of these innovations for clinical applications. Beyond short demos being shared on social media, there is little systematic examination for what the best use cases for production-grade clinical FMs are, or how healthcare organizations should weigh their benefits against their substantial risks [1,17–19]. Clinical FMs lack the shared evaluation frameworks and datasets [20] that have underpinned progress in other fields such as natural language processing (NLP) and computer vision [21]. This makes it difficult to quantify and compare these models' capabilities.

If we believe that FMs can help both providers and patients [22], then rigorous evaluations must be conducted to test these beliefs. In this review, we uncover notable limitations in how clinical FMs are evaluated and a large disconnect between their evaluation regimes and assumed clinical value. While adopting FMs into healthcare has immense potential [23], until we know how to evaluate whether these models are useful, fair, and reliable, it is difficult to justify their use in clinical practice. Inspired by recent efforts to holistically evaluate LLMs trained on non-clinical text for a range of capabilities beyond accuracy[24], we believe that a similar approach is necessary to tie evaluation of FMs writ large with use cases that matter in healthcare.

To clarify these challenges, we reviewed over 80 different clinical FMs built from electronic medical record (EMR) data. We included all models trained on structured (e.g. billing codes, demographics, lab values, medications) and unstructured (e.g. progress notes, radiology reports, other clinical text) EMR data, but explicitly excluded images, genetics, and wearables to manage the scope of this review. We refer to the combination of structured and unstructured EMR data (excluding images) as simply "EMR data" or "clinical data" [25]. We refer to FMs built on these forms of clinical data as "clinical foundation models" or "clinical FMs." Our primary contributions are:

1. *To our knowledge, we present the largest review of clinical FMs for structured and unstructured EMR data.* We organize these models into a simple taxonomy to clearly delineate their architectures, training data, capabilities, and public accessibility.
2. *We summarize the currently used evaluation frameworks for clinical FMs and identify their limitations.* We explain why current evaluation tasks provide little evidence for the purported benefits of FMs to a health system.
3. *We propose an improved framework for evaluating clinical FMs*. We advocate for metrics, tasks, and datasets that better capture the presumed value of clinical FMs.

We begin with a brief overview of clinical FMs and define their inputs, outputs, and capabilities in **Section 2**. In **Section 3**, we summarize the primary value propositions of FMs for health systems. We provide an overview of the training data behind clinical FMs in **Section 4.1**, examine current evaluation regimens and identify their limitations in **Section 4.2**, and propose a framework for improving these evaluations in **Section 4.3**. Finally, we discuss the promise of clinical FMs for solving a diverse range of healthcare problems in **Section 5**.

## 2. What are Clinical FMs?

There are two broad categories of foundation models built from electronic medical record data: *C*linical *La*nguage *M*odels (*CLaMs*) and *F*oundation models for *EMRs* (*FEMRs*).

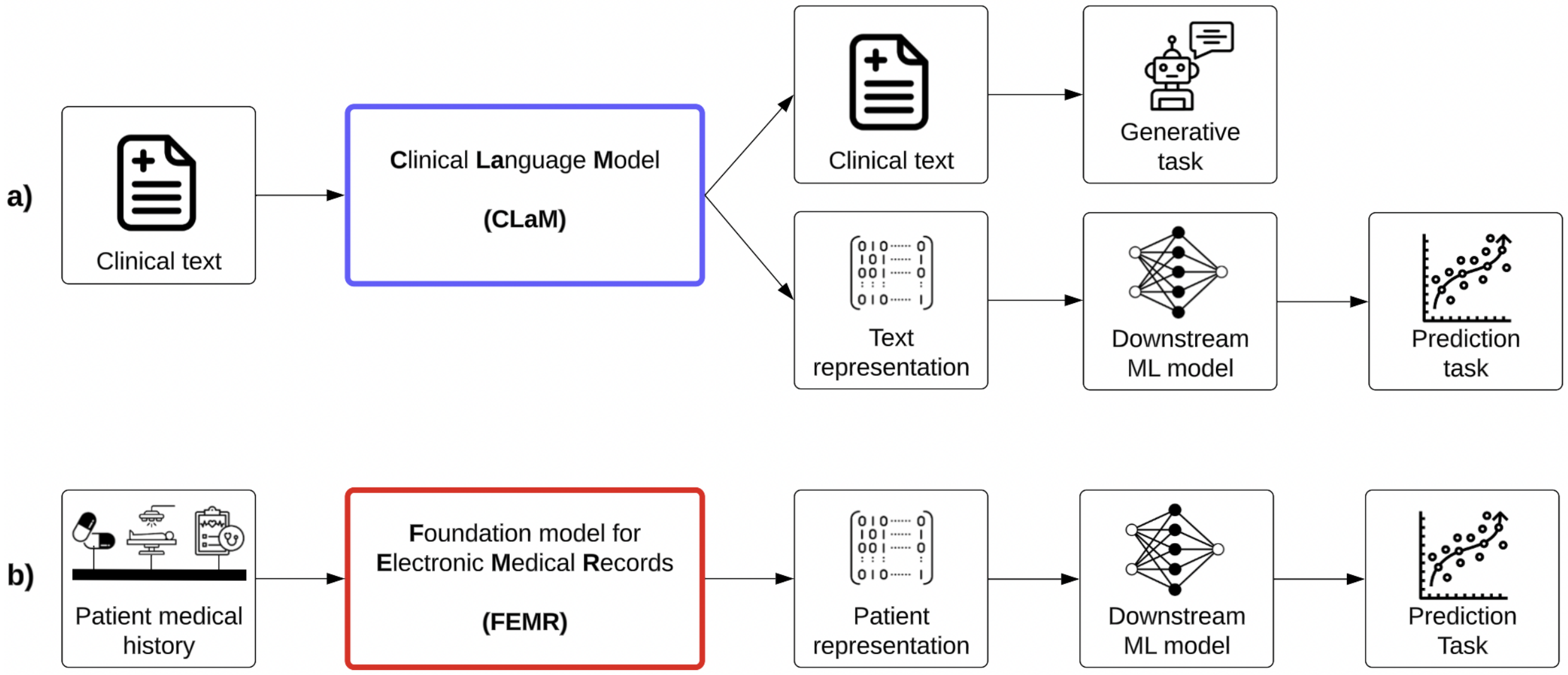

**Figure 1.** Overview of the inputs and outputs of the two main types of clinical FMs. (a) The inputs and outputs of Clinical Language Models (CLaMs). CLaMs ingest clinical text and output either clinical text or a machine-understandable representation of the input text which can then be used for downstream prediction tasks. (b) The inputs and outputs of Foundation models for Electronic Medical Records (FEMRs). FEMRs ingest a patient's medical history -- which is simply a sequence of medical events with some temporal ordering -- and output a machine-understandable representation of the patient which can then be used for downstream prediction tasks.

## Clinical Language Models (CLaMs)

The first category of FMs are *C*linical *La*nguage *M*odels, or *CLaMs*, which are a subtype of large language models (LLMs). As shown in **Figure 1a**, the unique attribute that separates CLaMs from general LLMs is their specialization on clinical/biomedical text – CLaMs are primarily trained on, ingest, and output clinical/biomedical text. For example, a CLaM could extract drug names from a doctor's note [26], automatically reply to patient questions [27], summarize medical dialogues [28], or predict mechanical ventilation needs based on clinical notes [29].

While general-purpose LLMs (e.g. ChatGPT, Bloom, GPT-3, etc.) trained on text scraped from the Internet can also be useful for clinical tasks, they tend to underperform CLaMs on domain-specific tasks [30,31], and thus we exclude them from this discussion. However, the conclusions from this review should also readily apply to these models.

## Foundation models for Electronic Medical Records (FEMRs)

The second class of clinical FMs are *F*oundation models for *E*lectronic *M*edical *R*ecords (*FEMRs*). These models are trained on the entire timeline of events in a patient's medical history. Given a patient's EMR as input, a FEMR will output not clinical text but rather a machine-understandable "representation" for that patient, as shown in **Figure 1b**. This representation – also referred to as a "patient embedding" – is typically a fixed-length, high-dimensional vector which condenses large amounts of patient information [32]. A patient's representation can then be used as input to any number of downstream models for different tasks. These downstream models (built on the "foundation" of FEMR representations) tend to be more accurate and robust than traditional machine learning (ML) models on clinically relevant tasks, such as predicting 30-day readmission or long length-of-stay [33].

The input to a FEMR can include many aspects of a patient's medical history, such as structured codes, lab values, claims, and clinical text. In practice, however, FEMRs are typically limited to the single modality of structured codes, as discussed in **Section 4.1**.

## 3. Benefits of Clinical FMs

Given the excitement around FMs in healthcare [23,34–39], we summarize their primary value propositions over traditional ML methods. These advantages could all be highly valuable to a health system. Thus, it is essential that our evaluation tasks, datasets, and metrics provide accurate assessments of these purported benefits.

1. ***Clinical FMs have better predictive performance.*** By using larger training datasets and more model parameters, FMs can achieve better predictive performance (e.g., higher sensitivity and specificity on classification tasks) than traditional ML models [32].
2. ***Clinical FMs require less labeled data ("improved sample efficiency").*** FMs enable superior model performance using fewer labeled data via "transfer learning" [40]. The core idea behind transfer learning is to first "pre-train" a model on large amounts of non-task-specific (and often unlabeled) data to teach the model general patterns. Then, the model is "fine-tuned" (i.e. continued to be trained) on a smaller dataset specific to the desired task. For example, a model pre-trained on the raw text of Wikipedia before being fine-tuned on a labeled dataset of 100 Tweets will outperform models solely trained on the smaller task-specific dataset of Tweets [40]. By learning representations that are useful for many downstream tasks via self-supervised pre-training, FMs reduce the need to label large, task-specific training datasets and therefore reduce the cost of building ML models for any particular task.
3. ***Clinical FMs enable simpler and cheaper model deployment.*** After a FM is trained, it can help to decrease the time, talent, and resources required to build subsequent ML models by serving as the figurative "foundation" upon which these subsequent applications are built [1]. Numerous companies have already commercialized this "ML-as-a-Service" approach, in which a centralized FM is made available to end-users via a simple API [41]. A similar approach could work in healthcare, wherein a clinical FM allows informaticians to integrate AI-related capabilities into applications while avoiding the expensive data ingestion, preprocessing, model training, and deployment steps in a typical ML pipeline [42].
4. ***Clinical FMs exhibit "emergent" capabilities that enable new clinical applications***. The large number of parameters in FMs has resulted in a phenomenon known as "emergence," in which previously intractable problems become tractable at sufficient model scale [7]. For example, CLaMs can now write coherent insurance appeals in ways thought impossible only a couple years ago [43], while FEMRs can generate compact patient representations that enable time-to-event modeling of hundreds of outcomes simultaneously [44].
5. ***Clinical FMs are more effective in handling multimodal data***. FMs can accept a wide range of data modalities (e.g. structured codes, lab values, clinical text, images, speech patterns, etc.) as inputs and incorporate them into a single unified representation [45]. This is especially useful in medicine given the many types of data produced by patients [46]. For example, a model might simultaneously consider an MRI scan, vital signs, and progress notes when predicting a patient's optimal treatment [47].

6. ***Clinical FMs offer novel interfaces for human-AI interaction.*** Via a technique called "prompting", a human can input natural language into an LLM and have the model respond in natural language [2]. This enables a two-way conversation between human and machine, and allows for the decomposition of problems into smaller steps via techniques such as "chain-of-thought" prompting [48]. Prompting generalizes beyond natural language. For example, a FEMR could be prompted with a desired clinical end state (e.g. normal A1C level) to identify which medications should be prescribed to achieve it [49].

## 4. State of Published Clinical FMs

We identified 84 distinct clinical FMs published before March 1, 2023. Specifically, we identified 50 CLaMs and 34 FEMRs, by following citations from a few seed publications and manual article curation. We focus exclusively on models that utilize structured and unstructured EMR data (excluding images) to scope this review.

### 4.1. Training Data and Public Availability of Models

**CLaMs**

***Training Data:*** CLaMs (**Figure 2a)** are primarily trained on either clinical text (i.e. documents written during the course of care delivery) or biomedical text (i.e. publications on biomedical topics). Almost all CLaMs trained on clinical text used a single database: MIMIC-III, which contains approximately 2 million notes written between 2001-2012 in the ICU of the Beth Israel Deaconess Medical Center [50]. CLaMs trained on biomedical text virtually always trained on PubMed abstracts and/or full-text articles. While most CLaMs trained on clinical text are also trained on biomedical text, the converse is not true.

***Model Availability:*** Almost all CLaMs have been made publicly accessible via online model repositories like HuggingFace [51]. Unfortunately, the exceptions are the very CLaMs that seem to have the best performance [52] -- ehrBERT [53], UCSF-Bert [52], and GatorTron [54] -- as they were trained on private EMR datasets.

***Takeaways****:* The high number of CLaMs published over the past several years may lead us to mistake motion for progress. Nearly all CLaMs have been trained on just two datasets -- MIMIC-III and PubMed, which respectively contain about 2 million clinical notes and 16 million abstracts with 5 million full-text publications. Combined, these two datasets contain about 18.5 billion words, which means models trained on them have substantial gaps in completeness (i.e. any scientific knowledge not contained within these corpora) and timeliness (i.e. any new diseases, treatments, or practices discovered after 2012 in the case of MIMIC-III).

**FEMRs**

***Training Data:*** Most FEMRs (**Figure 3a**) are trained on either small, publicly available EMR datasets or a single private health system's EMR database. Again, the most popular public dataset is MIMIC-III, which contains less than 40,000 patients [50]. Other public datasets vary greatly in size, from eICU's 139,000 patients [55] to the CPRD's longitudinal records on 7% of all patients in the UK [56]. Several FEMRs have been trained on insurance claims, which are typically larger in size and more diverse than EMR data but contain less granular information [57]. Examples of claims datasets include Truven Health MarketScan (170 million patients) [58] and Partners For Kids (1.8 million pediatric patients) [59]. In terms of data modalities, most FEMRs are unimodal as they only consider structured codes (e.g. LOINC, SNOMED, etc).

***Model Accessibility:*** FEMRs lack a common mechanism like HuggingFace for distributing models to the research community, as can be seen in the sparsity of the bottom-most row in **Figure 3a** compared to the density of the bottom-most row in **Figure 2a**. Few FEMRs have had their model weights published, meaning researchers must re-train these models from scratch on local EMR data to verify their performance.

***Takeaways:*** The overreliance on structured codes limits generalizability of FEMRs across health systems that use different EMR systems and coding practices. Some models, such as DescEmb, address this problem by first converting coded data into their textual descriptions, thus detaching the model from the specific codes on which it was trained [60]. An additional limitation of relying on coded data is that it contains inconsistencies and errors [61], and often provides an incomplete picture of patient state [62]. Some FEMRs have tackled this problem by combining unstructured EHR data (i.e. text) with structured EMR data to boost performance on specific phenotyping and prediction tasks [63,64]. However, the key unsolved challenge of how to publicly share pre-trained FEMRs continues to hinder the field's progress and precludes the primary value proposition of FMs -- namely, being able to build off a pre-trained model.

## 4.2. Current Evaluation of Clinical FMs

Clinical FMs are currently assessed on tasks that are relatively easy to evaluate. While these tasks provide diagnostic insights on model behavior, they provide limited insight on the claims of FMs being a "categorically different" technology [65,66], and offer little evidence for the clinical utility achieved by these models. Taking inspiration from the broader ML community's push towards Holistic Evaluation of Language Models [24], we do a critical evaluation of the evaluations currently used to evaluate clinical FMs.

### CLaMs

***Evaluation on Standard Tasks and Datasets.*** We collected every evaluation task that a CLaM was evaluated on in its original publication in **Figure 2b**, and grouped these tasks as they are commonly reported in the literature. Most CLaMs are being evaluated on traditional NLP-style tasks such as named entity recognition, relation extraction, and document classification on either MIMIC-III (clinical text) or PubMed (biomedical text) [67,68]. Given that clinical text has its own unique structure, grammar, abbreviations, terminology, formatting, and other idiosyncrasies not found in other domains [69], it is alarming that roughly half of all CLaMs surveyed were not validated on clinical text, and thus may be overestimating their expected performance in a healthcare setting.

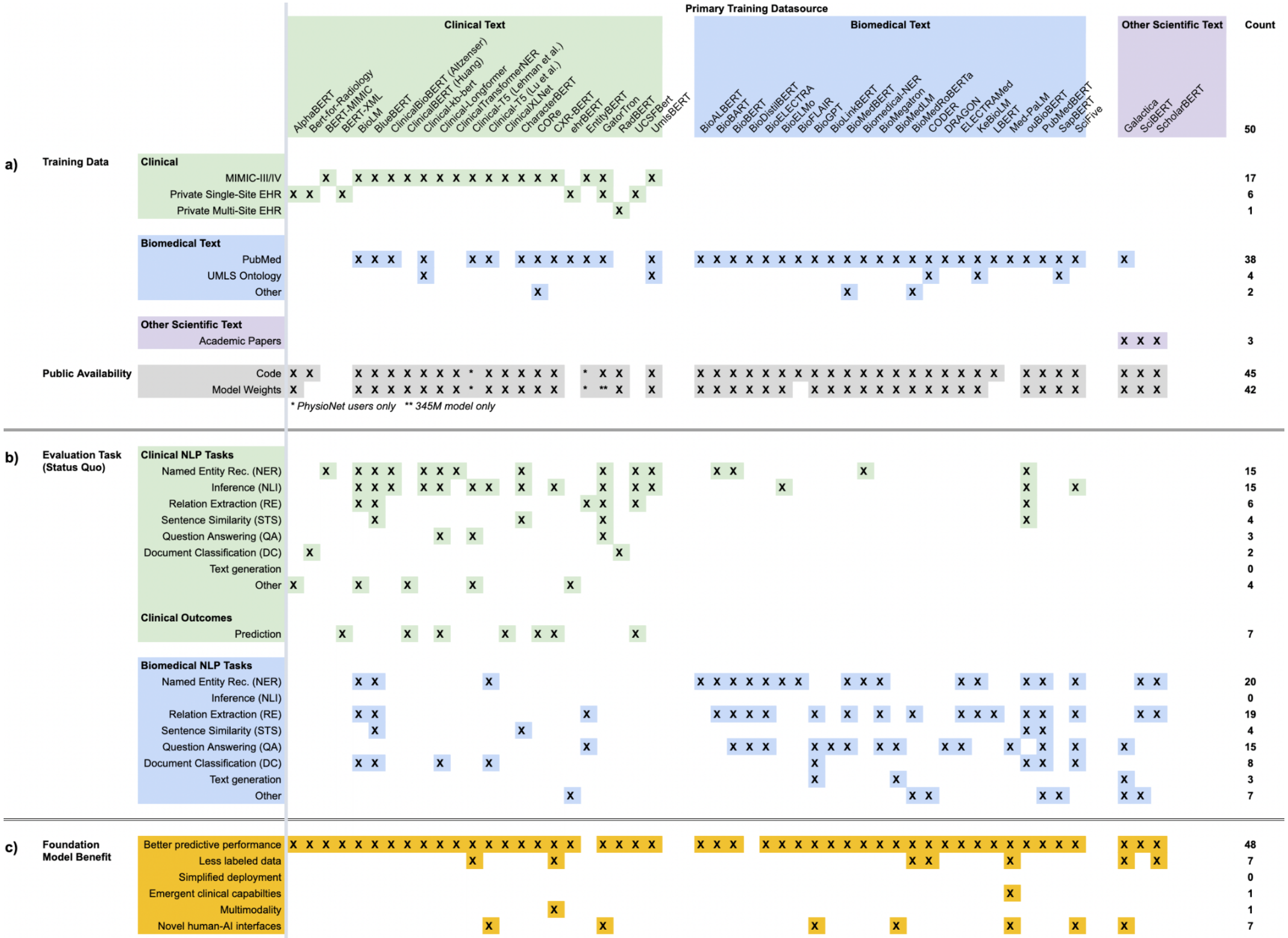

**Figure 2.** A summary of CLaMs and how they were trained, evaluated, and published. Each column is a specific CLaM, grouped by the primary type of data they were trained on. Columnwise, the CLaMs primarily trained on clinical text are green (n=23), those trained primarily on biomedical text are blue (n=24), and models trained on general academic text are purple (n=3).
**(a)** Training data and public availability of each model. The top rows mark whether a CLaM was trained on a specific dataset, while the bottom-most row records whether a model's code and weights have been published. Almost all CLaMs have had their model weights published, typically via shared repositories like the HuggingFace Model Hub.
**(b)** Evaluation tasks on which each model was evaluated in its original paper. Green rows are tasks whose data was sourced from clinical text and blue rows are evaluation tasks sourced from biomedical text. The tasks are presented by the way they are commonly organized in the literature. CLaMs primarily trained on clinical text are evaluated on tasks drawn from clinical datasets, while CLaMs primarily trained on biomedical text are almost exclusively evaluated on tasks that contain general biomedical text (i.e. not clinical text). **(c)** Clinical FM benefits on which each model was evaluated in its original paper. The underlying tasks presented in this section are identical to those in **(b),** but here the tasks are re-organized into six buckets that reflect the six primary FM benefits described in **Section 3**. While almost all CLaMs have demonstrated the ability to improve predictive accuracy over traditional ML approaches, there is scant evidence for the other five value propositions of clinical FMs.

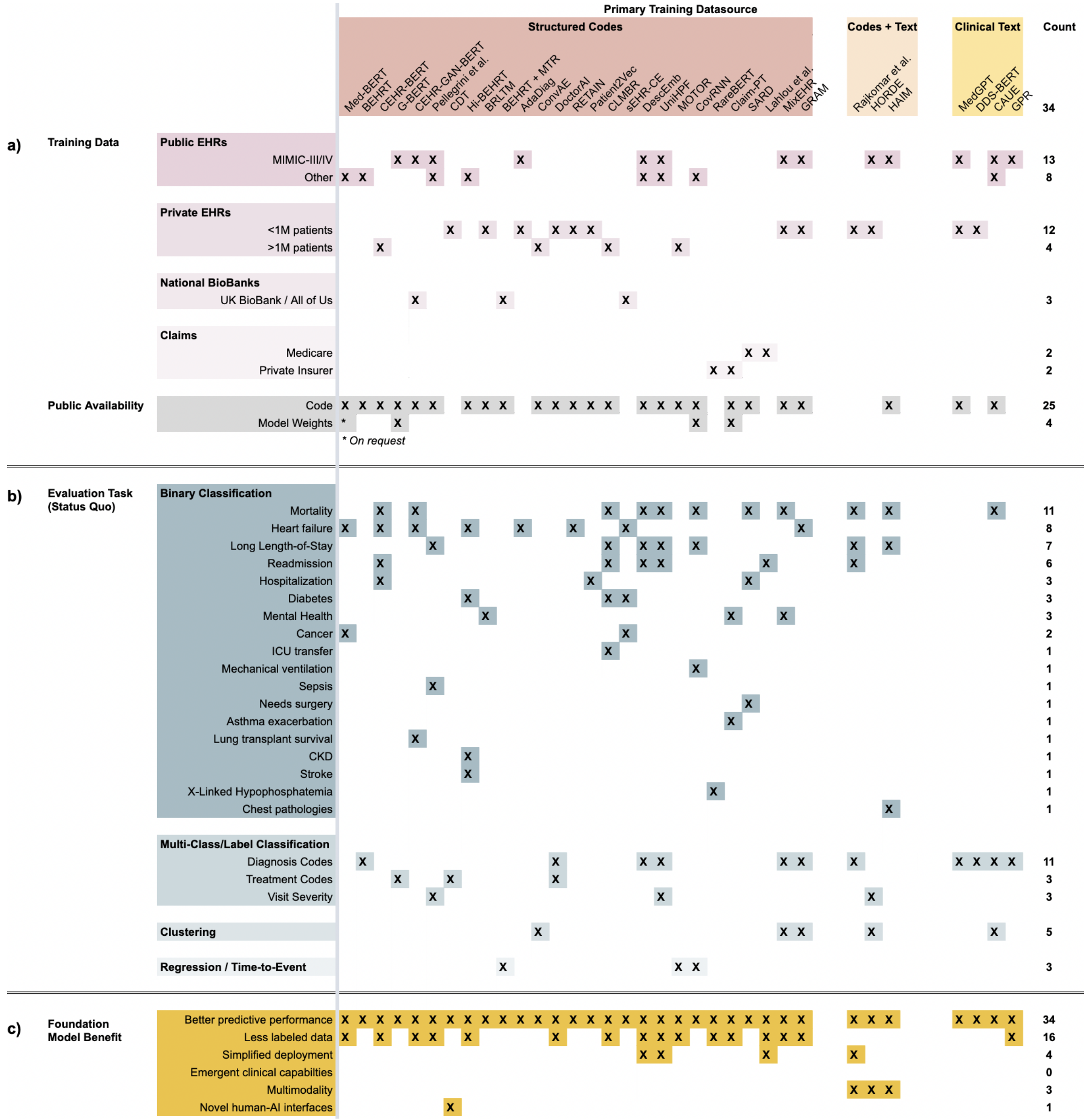

**Figure 3.** A summary of FEMRs and how they were trained, evaluated, and published. Each column is a specific FEMR, grouped by the primary type of data they were trained on. Columnwise, the FEMRs primarily trained on structured EMR codes (e.g. billing, medications, etc.) are red (n=27), those trained on both structured codes and clinical text are orange (n=3), and models trained only on clinical text are yellow (n=4).

**(a)** Training data and public availability of each model. The top rows mark whether a FEMR was trained on a specific dataset, while the bottom-most row records whether a model's code and weights have been published. Very few FEMRs have had their model weights published, as they are limited by data privacy concerns and a lack of interoperability between EMR schemas.

**(b)** Evaluation tasks on which each model was evaluated in its original paper. From top to bottom, the evaluation tasks are binary classification, multi-class/label classification, clustering of patients/diseases, and regression tasks like time-to-event. The tasks are presented by the way they are commonly organized in the literature. FEMRs are evaluated on a very broad and sparse set of evaluation tasks -- even the same nominal task will often have different definitions across papers.

**(c)** Clinical FM benefits on which each model was evaluated in its original paper. The underlying tasks presented in this section are identical to those in **(b),** but here the tasks are re-organized into six buckets that reflect the six primary FM benefits described in **Section 3**. While almost all FEMRs have demonstrated the ability to improve predictive accuracy over traditional ML approaches, and a significant number have demonstrated improved sample efficiency, there is scant evidence for the other four value propositions of clinical FMs.

When NLP tasks are sourced from clinical text, they can be useful measures of a model's linguistic capabilities. However, these NLP tasks are greatly limited by their overreliance on the same handful of data sources [68], small dataset sizes (typically thousands of examples) [68,70], highly repetitive content [71], and low coverage of use cases [20]. As a result, strong performance on a clinical NLP task does not provide compelling evidence to a hospital looking to deploy a CLaM -- claiming that *"Model A achieves high precision on named entity recognition on 2,000 discharge notes from MIMIC-III*" is very different than "*Model A should be deployed across all of Health System X to identify patients at risk of suicide*".

***Evaluation on FM Benefits.*** To illustrate the disconnect between current evaluation tasks and the loftier promises of clinical FMs, we reorganized the rows of evaluation tasks from **Figure 2b** -- originally presented as they are typically grouped in the literature -- along the six primary FM value propositions from **Section 3.** The result is **Figure 2c**, which identifies which CLaMs were evaluated against any of the six core benefits of clinical FMs. Most CLaMs have only shown evidence for one FM value proposition: improved predictive accuracy on certain tasks. However, there is little evidence supporting the other purported benefits of FMs, such as simplified model deployment or reducing the need for labeled data. In other words, there is a gap in our understanding of what CLaMs *can do* versus what CLaMs can do *that is valuable to a health system and which traditional ML models cannot do*.

**FEMRs**

***Evaluation on Standard Tasks and Datasets.*** We collected the original tasks on which each FEMR was evaluated in **Figure 3b** and bucketed them as they are typically presented in the literature. Evaluation of FEMRs is in an even poorer state than that of CLaMs. While CLaMs benefit from the NLP community's adoption of standardized task formats, FEMRs lack a similar set of "canonical" evaluations. Instead, FEMRs are evaluated on an extremely sparse set of tasks with little-to-no overlap across publications. This makes it highly non-trivial to compare the performance of different FEMRs.

These tasks are typically grouped by how each task is formulated, e.g. binary classification v. multi-label classification v. regression. The most popular prediction tasks are binary classification tasks such as mortality, heart failure, and long length-of-stay, but even the same nominal task can have widely divergent definitions across papers [72].

***Evaluation on FM Benefits.*** We reorganized the rows of evaluation tasks from **Figure 3b** along the six primary value propositions of clinical FMs listed in **Section 3.** The result is **Figure 3c,** which shows that almost all evaluations of FEMRs have been focused on demonstrating their superior predictive accuracy over traditional ML models. Notably, the ability to use less labeled data (i.e. sample efficiency) has been fairly well-documented with FEMRs. However, the other four potential benefits of FMs have gone largely unstudied. And while evaluations of predictive accuracy are straightforward to perform, it is not the sole property of FMs that would justify their adoption by a health system.

## 4.3. Improved Evaluation Paradigms for Clinical FMs

To better quantify the ability of clinical FMs to achieve the six key benefits of FMs outlined in **Section 3,** we propose several improved evaluation metrics and tasks in **Figure 4**. Our suggestions are by no means comprehensive, but rather meant to spark further discussion on how to align model evaluation with demonstration of clinical value.

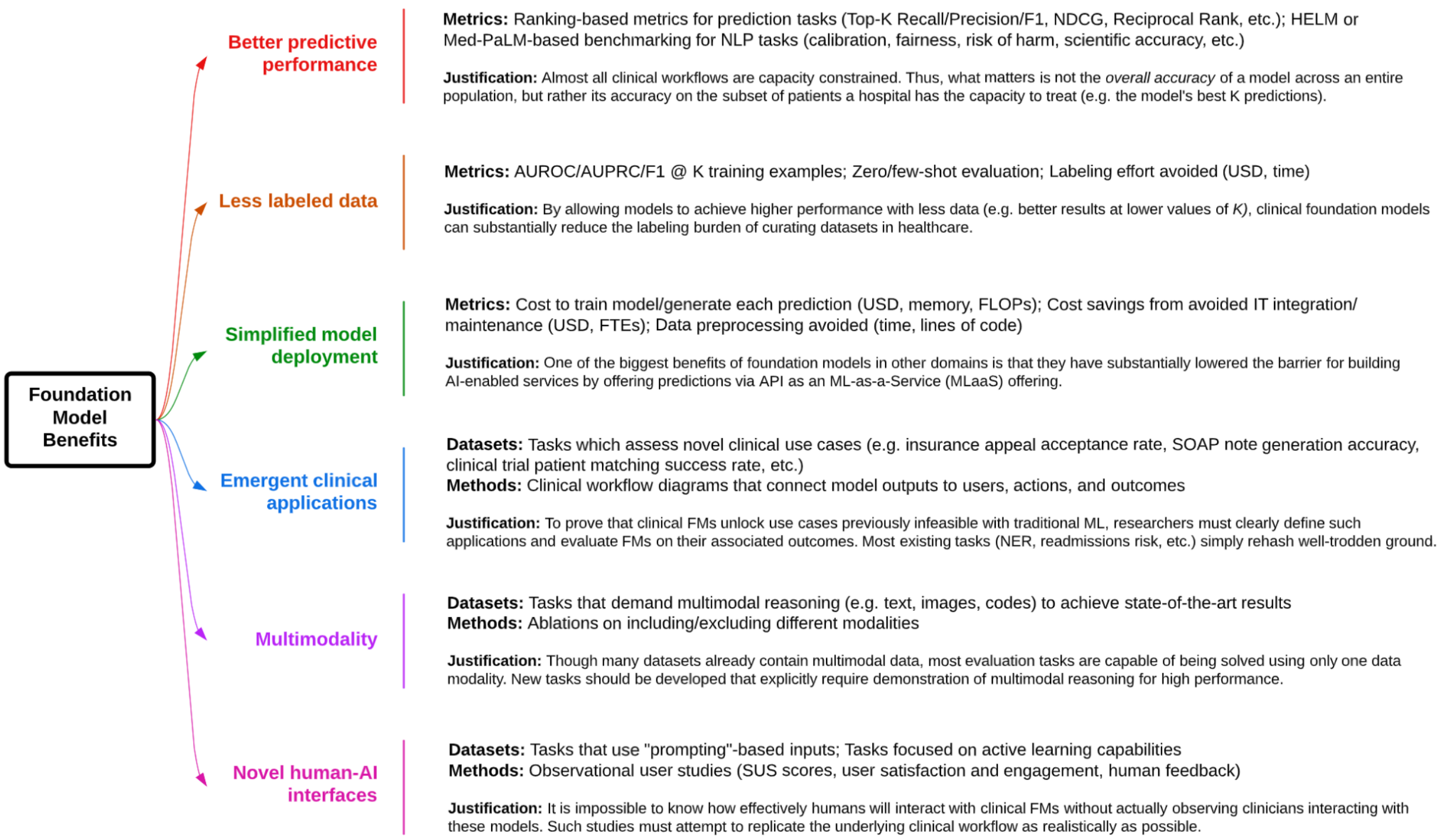

**Figure 4.** Proposals for how to demonstrate the value of CLaMs and FEMRs for achieving the six primary value propositions of FMs to health systems over traditional ML models.

## 1. Better Predictive Performance

The most thoroughly studied property of clinical FMs has been their improved predictive performance on classification and regression tasks based on AUROC, AUPRC, F1 Score, and Accuracy. These metrics assume infinite capacity to act on a model's predictions. In reality, clinical workflows are capacity constrained – a nursing team may only be able to act on a handful of model predictions per day [73,74]. Thus, a health system should only care about a model's accuracy on patients for which it has the capacity to intervene. We therefore recommend that researchers adopt ranking-based metrics (e.g. Top-K precision/recall/F1, reciprocal ranking, etc.), which are commonly used for recommendation systems [75]. Additionally, we propose examining not just a model's ability to classify patients correctly but also its calibration across subgroups, fairness, and alignment with clinical best practices [24,27]. Human evaluation may be necessary in some cases, such as evaluating a CLaM's ability to accurately write answers to clinical questions [27].

## 2. Less Labeled Data

The simplest way for researchers to demonstrate how clinical FMs exhibit improved sample efficiency is to replace evaluation metric "*X*" with the more nuanced metric "*X using K training examples*". For example, replacing "*AUROC*" with "*AUROC using 1,000 labeled radiology reports for fine-tuning.*" Ideally, a clinical FM would enable similar model performance at low values of *K* as at high values of *K*. Another way to demonstrate improved sample efficiency is to measure *zero-shot* and *few-shot* model performance, in which a model is given either zero or <100 examples, respectively, for the task on which it is evaluated. One could also measure the *total dataset annotation time saved* by using a clinical FM, measured in terms of dollars or hours.

### 3. Simplified Model Deployment

To quantify the value of FMs in lowering the barrier for building task-specific models [1,76], one possible metric is the cost of hardware/compute/memory needed to train a model or generate a prediction. More broadly, we can measure the overall cost savings of using a clinical FM in terms of full-time equivalents (FTEs) or resource hours saved when downstream models (e.g. risk of inpatient mortality) are built on top of a clinical FM versus training a task-specific model from scratch. We recognize, however, that this evaluation may be the most challenging to conduct, as it requires buy-in from the business, clinical, and IT units of a health system. Health systems with dedicated ML Operations ("MLOps") teams may be better positioned to realize these benefits [77].

### 4. Emergent Clinical Applications

Clinical FMs can also perform entirely new tasks thought to be beyond the reach of machines even just a year ago, e.g. summarizing MRI reports in patient accessible terms, writing discharge instructions, or generating differential diagnoses [43,78]. To prove that success in these tasks goes beyond cherry-picked demos, researchers must articulate how their models fit into clinical workflows. This will require collaboration with clinicians and informaticists to tie model outputs to specific actions taken by care team members [79]. Instead of hand-waving towards a future of AI doctors, we must explicitly define the scenarios in which clinical FMs should first get incorporated and measure user adherence and adoption.

### 5. Multimodality

Currently, the majority of evaluation tasks span one data modality [72], even though models that simultaneously use multiple data modalities show substantial gains [80]. There is a strong unmet need for evaluation scenarios which explicitly require multimodal representations. Many datasets already include multimodal data (e.g. MIMIC-III, eICU, private EMRs, etc.), but evaluation tasks are not constructed in ways that require demonstration of multimodal reasoning. A great example of datasets that accomplish this are the Holistic AI Framework (HAIM), which builds on top of MIMIC-III to enable truly multimodal evaluation scenarios [81], as well as CheXpert and MS-CXR for paired radiology images and text [82].

### 6. Novel Human-AI Interfaces

Human evaluation and usability studies are needed to quantify the utility of interacting with FMs via prompts [1]. Metrics include user satisfaction, engagement, system usability scale scores, qualitative interview feedback, and the time/effort required to achieve stated goals [83–85]. Measuring the skill-level necessary to operate a model can also shed light on its ability to empower providers to perform a multitude of roles. For FEMRs, an accepted paradigm for "prompting" does not yet exist, so developing a framework for prompting a patient's medical history would represent a significant step forward. One exception is the Clinical Decision Transformer, which used a desired clinical end state (e.g. normal A1C levels) as a prompt to generate medication recommendations [49].

## 5. Discussion

Our review of 50 CLaMs and 34 FEMRs shows that most clinical FMs are being evaluated on tasks that provide little information on the potential advantages of FMs over traditional ML models. While there is ample evidence that clinical FMs enable more accurate model predictions, **Figure 2** and **Figure 3** show that minimal work has been conducted to validate whether the other, potentially more valuable benefits of FMs will be realized in healthcare. These benefits include reducing the burden of labeling data, offering novel human-AI interfaces, and enabling new clinical applications beyond the reach of traditional ML models, among others outlined in **Section 3**. To help bridge this divide, we advocate for the development of new evaluation tasks, metrics, and datasets more directly tied to clinical utility, as summarized in **Figure 4**.

While we focused this review on the potential benefits of clinical FMs, these technologies present numerous risks that must also be considered and investigated. Data privacy and security are significant concerns with FMs, as they may leak protected health information through model weights or prompt injection attacks [86,87]. FMs are also more difficult to interpret, edit, and control due to their immense size [88]. They require high up-front costs to create, and while these costs can be amortized over multiple downstream applications, their value may take longer to realize than a smaller model developed for a single high-value task [89]. Additionally, FMs may fall under Software-as-a-Medical-Device guidelines regulating their usage in the clinic [90]. And similar to traditional ML models, FMs are susceptible to biases induced by mis-calibration or overfitting [91], as well as inducing “automation bias” in which clinicians defer to a model’s outputs even when they are obviously incorrect [92]. Developing frameworks for determining a model’s overall worth remain indispensable [73].

Despite these challenges, clinical FMs hold immense promise for solving a diverse range of healthcare problems. We invite the research community to develop better evaluations to help realize their potential for benefiting both patients and providers [22].

## 6. Acknowledgements

MW is supported by an NSF Graduate Research Fellowship. MW, YX, RT, JF, ES, SF, MP, and NHS also acknowledge support from Stanford Medicine for this research.

## 7. Competing Interests

BP reports stock-based compensation from Google, LLC. Otherwise, the authors declare that there are no competing interests.

## 8. Author Contributions

MW, YX, JF, and NHS conceptualized and designed the study; MW, YX, and RT extracted data; MW, YX, BP, RT, JF, and NHS conducted the analysis and wrote the manuscript. MW, YX, BP, RT, JF, NHS, and MP revised the manuscript. ES and SF contributed to the analysis. All authors approved the final version of the manuscript and take accountability for all aspects of the work.